\documentclass{article}

\usepackage{arxiv}

\usepackage[utf8]{inputenc} 
\usepackage[T1]{fontenc}    
\usepackage{hyperref}       
\usepackage{url}            
\usepackage{booktabs}       
\usepackage{amsfonts}       
\usepackage{nicefrac}       
\usepackage{microtype}      
\usepackage{lipsum}
\usepackage[linesnumbered,ruled,vlined]{algorithm2e}
\usepackage{graphicx}
\usepackage{amsmath}
\usepackage{authblk}

\title{ss-Mamba: Semantic-Spline Selective State-Space Model}

\author{Zuochen Ye}

\affil{Department of Statistics \\
        National Chengchi University \\
        \texttt{112354016@nccu.edu.tw}}

\begin{document}
\maketitle
\begin{abstract}
We propose ss-Mamba, a novel foundation model that enhances time-series forecasting by integrating semantic-aware embeddings and adaptive spline-based temporal encoding within a selective state-space modeling framework. Building upon the recent success of Transformer architectures, ss-Mamba adopts the Mamba selective state-space model as an efficient alternative that achieves comparable performance while significantly reducing computational complexity from quadratic to linear time. Semantic index embeddings, initialized from pretrained language models, allow effective generalization to previously unseen series through meaningful semantic priors. Additionally, spline-based Kolmogorov–Arnold Networks (KAN) dynamically and interpretably capture complex seasonalities and non-stationary temporal effects, providing a powerful enhancement over conventional temporal feature encodings. Extensive experimental evaluations confirm that ss-Mamba delivers superior accuracy, robustness, and interpretability, demonstrating its capability as a versatile and computationally efficient alternative to traditional Transformer-based models in time-series forecasting.
\end{abstract}


\section{Introduction}
\label{sec:intro}

Time-series data permeate numerous domains, including financial markets, climate science, and manufacturing, where precise forecasting significantly influences strategic decisions. Traditional forecasting methods are predominantly task-specific or series-specific, restricting their ability to generalize effectively across diverse and heterogeneous datasets. Inspired by the transformative success of large language models (LLMs) in cross-task transfer learning and few-shot generalization, recent advancements such as Nixtla's TimeGPT\cite{NixtlaTimeGPT} and Google's TimesFM\cite{GoogleTimesFM} underscore the potential for foundational models capable of robust zero-shot forecasting for previously unseen series.

In response to these limitations and opportunities, we introduce \textbf{ss-Mamba}, a groundbreaking multimodal time-series foundation model explicitly designed to leverage multimodal contextual embeddings for enhanced cross-series forecasting. Our proposed model synergistically integrates semantic index embeddings, spline-based temporal feature encodings via Kolmogorov–Arnold Networks (KAN)\cite{Kan2025}, and the highly efficient selective state-space backbone of Mamba\cite{Gu2025Mamba}.

By leveraging semantic embeddings from pretrained language models like BERT\cite{Devlin2019}, \textbf{ss-Mamba} harnesses the semantic richness inherent in series identifiers, effectively distinguishing contextually meaningful differences between series (e.g., `Gold Price' vs. `Silver Price'). The spline-based KAN module further complements this by offering flexible, interpretable representations of temporal features, efficiently capturing complex seasonal and non-linear patterns without excessive parameterization.

Central to our approach is the Mamba architecture, recognized for its linear-time complexity and exceptional computational efficiency, achieving effective modeling of long-term dependencies that rival the quadratic complexity traditionally associated with Transformers\cite{Vaswani2017}. Unlike conventional SSMs and Transformer-based models, our model incorporates an innovative selection mechanism, enabling adaptive attention and selective retention of salient inputs. This ensures efficient contextual understanding, dynamically prioritizing essential temporal and semantic information.

This fusion of multimodal semantic knowledge and state-of-the-art temporal modeling significantly expands generalization capabilities. \textbf{ss-Mamba} demonstrates substantial potential in zero-shot scenarios, effectively predicting series not previously encountered during training. Furthermore, by implementing a GPU-optimized framework, incorporating mixed-precision training \cite{Micikevicius2018} and gradient clipping \cite{Pascanu2013}, our model maintains computational efficiency and stability, making it suitable for real-world, large-scale applications.

To rigorously evaluate our proposed approach, we conduct extensive empirical assessments, including single-series and multi-series tasks, zero-shot predictions for novel series, and comprehensive ablation studies to quantify the contribution of each component. The results affirm that \textbf{ss-Mamba} not only delivers exceptional predictive accuracy but also considerably advances foundational forecasting models' capacity to generalize robustly across diverse and multimodal series. This positions \textbf{ss-Mamba} at the forefront of innovative developments in multimodal time-series forecasting, marking a substantial advancement over existing state-of-the-art methodologies.

\section{Literature Review}
\label{sec:review}

Time-series forecasting (TSF) remains a critical research area given its vast applications in domains like financial markets, climate sciences, and manufacturing. Traditional TSF models often focus on single-series or single-task predictions, limiting their adaptability to diverse, cross-domain datasets, and reducing their overall generalizability \cite{Liu2024}. The recent proliferation of large-language models (LLMs) has spurred interest in importing the foundation model paradigm to TSF, whereby one pre-trained model can effectively perform forecasting, classification, and anomaly detection tasks simultaneously \cite{GoogleTimesFM, Goswami2024, Ye2024}.

Incorporating semantic knowledge about each series through embeddings is one promising avenue for enhancing the cross-domain capabilities of TSF models. This approach involves initializing trainable embedding matrices with vectors derived from pre-trained language models such as BERT or lightweight models like Word2Vec \cite{Kazemi2019}. Such semantic embeddings enable the model to generalize effectively by leveraging prior lexical knowledge, allowing the network to intuitively borrow patterns from semantically similar series \cite{Kazemi2019}. This method notably supports zero-shot forecasting capabilities, where new series can be effectively predicted without prior direct training, simply through their semantic embedding representations \cite{Goswami2024}.

A significant advancement in time-feature encoding involves the use of Kolmogorov–Arnold Networks (KAN), inspired by the Kolmogorov–Arnold representation theorem. KAN leverages learnable spline functions rather than traditional linear weights, allowing networks to flexibly approximate any univariate transformations per input dimension \cite{Xu202406}. This method proves particularly advantageous over traditional fixed encodings, such as sine and cosine transformations, or even vector-based representations like Time2Vec, by dynamically adapting to non-stationary or multi-period patterns and offering interpretability through exportable analytic formulas \cite{Xu202406, Kazemi2019}.

Sequential modeling, critical for capturing temporal dependencies within time series, has seen substantial innovation through models such as Mamba Selective State-Space Models (SSM). Unlike conventional Transformer architectures with quadratic complexity, Mamba leverages a linear-time selective SSM approach that dynamically updates hidden states based on content \cite{Gu2025Mamba}. This content-dependent selective state-space approach provides attention-like capabilities without the substantial computational costs typically associated with traditional Transformers, making it highly efficient for long sequences \cite{Karadag2025, Gu2025Mamba}. Mamba models have achieved state-of-the-art accuracy on extensive long-sequence forecasting tasks, often requiring fewer parameters and offering significantly improved inference speeds \cite{Karadag2025}.

In conclusion, the fusion of semantic embeddings, spline-based temporal encoding via KAN, and efficient sequence modeling with Mamba represents a compelling direction for TSF. This integrated approach promises substantial enhancements in model interpretability, scalability, and generalization capability, setting a robust foundation for future TSF research and applications \cite{Xu202405, Gu2025Mamba, Karadag2025}.

\section{Proposed Method: ss-Mamba}
\label{sec:method}

Our goal is to build a \emph{daily-frequency foundation model} that can forecast \emph{heterogeneous} time-series with a \textbf{single} set of parameters.  The architecture (Fig.~\ref{fig:ss-Mamba-model}) couples three learnable blocks:

\begin{figure}
    \centering
    \includegraphics[width=0.4\linewidth]{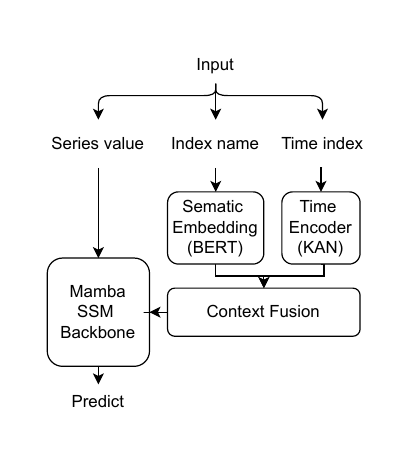}
    \caption{ss-Mamba model architecture}
    \label{fig:ss-Mamba-model}
\end{figure}

\begin{enumerate}
  \item \textbf{Semantic Index Embedding} $\mathbf{e}^{(\text{idx})}$ injects
        textual knowledge about each series (e.\,g.\ “S\&P 500” versus
        “Taipei Temperature”).
  \item \textbf{Spline-based Temporal Encoder} $\mathbf{e}^{(\text{time})}$
        maps raw calendar features to a dense vector via a
        Kolmogorov–Arnold Network (KAN), capturing flexible seasonality and
        trend effects.
  \item \textbf{Mamba Selective State-Space Backbone} models sequence dynamics
        with linear time/space complexity while being conditioned on the two
        embeddings above.
\end{enumerate}

Together they form the \emph{ss-Mamba} network (semantic–spline Mamba).

\subsection{Semantic Index Embedding}
\label{subsec:index-embed}

A foundational aspect of our framework is the explicit modeling of each time series' static properties and domain semantics. We accomplish this by encoding the identifier of each series—typically an arbitrary descriptive name—into a dense, information-rich vector. This vector serves as a persistent representation within the model's \emph{index space}, acting as a memory cell for all knowledge specific to the series and facilitating knowledge transfer across related series.

\paragraph{Language Model Index Space.}
Let $\mathcal{S}$ denote the set of all unique time series in the dataset. For each series $s \in \mathcal{S}$, we begin by mapping its natural-language identifier $n_s$ into a contextual semantic vector using a pre-trained language model:

\begin{equation}
\mathbf{h}^{(\text{BERT})}_s = f_{\mathrm{BERT}}(n_s) \in \mathbb{R}^{d_{\mathrm{BERT}}},
\end{equation}

where $f_{\mathrm{BERT}}(\cdot)$ extracts the final hidden state (typically the \texttt{[CLS]} embedding) from the language model, and $d_{\mathrm{BERT}}$ is the output dimension of BERT’s last layer (e.g., 768 in standard implementations, not shown in the equation).

To align the semantic vector with the internal state dimension of the downstream model, we project it into the required index embedding space using a trainable linear transformation:

\begin{equation}
\mathbf{e}_s = \mathbf{W}_\mathrm{proj} , \mathbf{h}^{(\text{BERT})}_s + \mathbf{b}_\mathrm{proj}, \qquad \mathbf{W}\mathrm{proj} \in \mathbb{R}^{N \times d_{\mathrm{BERT}}},
\end{equation}

where $N$ is the hidden size used throughout the Mamba state-space model, and $\mathbf{e}_s \in \mathbb{R}^N$ is the semantic index embedding for series $s$.

\paragraph{Index Space as Memory.}
This two-step process—semantic encoding and linear projection—constructs a continuous \emph{index space} in which each series is mapped to a unique vector. All parameters are jointly optimized during training. The language model ensures that series with similar semantic attributes (e.g., 'gold price' and 'silver price') are initially embedded close together, while the subsequent fine-tuning specializes these vectors for the forecasting task.

Formally, the complete semantic embedding module can be viewed as a mapping:

\begin{equation}
\mathbf{e}_s = f_{\mathrm{IDX}}(n_s) \in \mathbb{R}^N,
\end{equation}

where $f_{\mathrm{IDX}}(\cdot) = \mathbf{W}_\mathrm{proj} , f_{\mathrm{BERT}}(\cdot) + \mathbf{b}_\mathrm{proj}$, collectively defining the “index space” that stores the static information for each series.

\paragraph{Generalization and Zero-shot Capability.}
Because this embedding pipeline is constructed from a pre-trained language model, the system is naturally capable of encoding unseen or novel series identifiers at inference time: by applying the same $f_{\mathrm{IDX}}(\cdot)$ mapping, the model can retrieve meaningful representations for new series with minimal or no task-specific retraining.

\paragraph{Summary.}
In summary, the semantic index embedding module forms a memory-augmented “index space” that preserves all relevant series information in a vector format. These embeddings are fine-tuned with the end-to-end forecasting objective, bridging static linguistic priors and dynamic numerical modeling in a unified architecture.

All index embeddings are updated via gradient descent during training. At inference, a new series can be embedded \emph{zero-shot} by applying the same pipeline to its name, providing immediate generalization capability to novel or out-of-domain series.

\subsection{Temporal Encoding via Kolmogorov–Arnold Spline Networks}
\label{subsec:kan}

Capturing the intricate effects of calendar time—such as seasonality, business cycles, and holiday shifts—is essential for accurate time series modeling. Rather than relying on fixed periodic bases, we introduce a learnable and highly expressive temporal encoder based on Kolmogorov–Arnold Networks (KANs). This approach allows the model to discover arbitrary and potentially non-periodic patterns in the temporal structure of the data.

\paragraph{Calendar Descriptor Vector.}
For each time step $t$, we construct a raw calendar feature vector:
\begin{equation}
\mathbf{d}_t = \bigl[
\mathrm{ordinal}_t,
\mathrm{year}_t,
\mathrm{month}_t,
\mathrm{day}_t,
\mathrm{dow}_t,
\mathrm{doy}_t,
\mathrm{quarter}_t
\bigr]^{\top} \in \mathbb{R}^{k},
\end{equation}
where $k$ is the number of extracted calendar features.

\paragraph{Spline-based Univariate Transformations.}
Each component $d_{t,j}$ of $\mathbf{d}_t$ is passed through an independently parameterized, smooth univariate function implemented as a B-spline expansion:

\begin{equation}
g_j(x) = \sum_{r=1}^{R} \alpha_{j, r} B_{r, m}(x; \boldsymbol{\xi}_j),
\end{equation}

where $B_{r, m}(\cdot; \boldsymbol{\xi}_j)$ denotes the $r$-th B-spline of degree $m$ with knot vector $\boldsymbol{\xi}_j$, and $\alpha_{j, r}$ are trainable coefficients. The spline parameters are optimized jointly with the rest of the network. The outputs are collected as:

\begin{equation}
u_{t, j} = g_j(d_{t, j}), \qquad \mathbf{u}_t = [u_{t, 1}, \ldots, u_{t, k}]^\top \in \mathbb{R}^k.
\end{equation}

\paragraph{Linear Mixing and Temporal Encoding Vector.}
To enable flexible combinations of these temporal basis functions, the spline outputs are linearly mixed and transformed:
\begin{equation}
\mathbf{z}^{(\mathrm{TEV})}_t = \sigma(\mathbf{W} \mathbf{u}_t + \mathbf{b}) \in \mathbb{R}^{N},
\end{equation}

where $\mathbf{W} \in \mathbb{R}^{N \times k}$ and $\mathbf{b} \in \mathbb{R}^{N}$ are learnable parameters, $N$ is the model’s internal dimension, and $\sigma$ is an activation function such as $\tanh$ or the identity.

\paragraph{KAN as Universal Time Encoder.}
The Kolmogorov–Arnold theorem ensures that such a construction is theoretically capable of approximating any continuous function of the calendar features, provided sufficient spline resolution and hidden dimension. In practice, modest settings (e.g., $m \leq 3$, $R \leq 16$, and $N$ between 64 and 128) suffice to model complex effects including variable-length cycles, abrupt shifts, and irregular seasonalities.

\paragraph{Integration and Generalization.}
The resulting Temporal Encoding Vector (TEV) $\mathbf{z}^{(\mathrm{TEV})}_t$ is a compact, learnable fingerprint of time that is fed into the model as part of its context representation. Unlike fixed trigonometric encodings, KAN-based TEVs adapt flexibly to arbitrary calendrical phenomena, supporting robust generalization across domains and time horizons.

\begin{equation}
\mathbf{k}_t = \mathbf{z}^{(\mathrm{TEV})}_t \in \mathbb{R}^{N}
\end{equation}

All parameters are updated via gradient descent along with the forecasting objective, ensuring that temporal features are optimally encoded for the downstream prediction task.

\subsection{Selective-State-Space Backbone (Mamba) with Contextualized Input Matrix}
\label{subsec:mamba}

The core of our architecture is a selective state-space sequence model inspired by the recent Mamba framework. This backbone combines the efficiency of recurrent state updates with content-aware gating, enabling long-term memory and flexible context integration for time-series forecasting.

\paragraph{Original State-Space Recurrence.}
At each position $t$, the backbone maintains a hidden state $\mathbf{h}_t \in \mathbb{R}^N$, which evolves according to a parameterized linear recurrence:
\begin{equation}
\mathbf{h}_{t+1} = \bar{\mathbf{A}}_t \mathbf{h}_t + \bar{\mathbf{B}}_t x_t,
\end{equation}
where $\bar{\mathbf{A}}_t, \bar{\mathbf{B}}_t \in \mathbb{R}^N$ are the discretized transition and input matrices, and $x_t$ is the (potentially projected) input value at time $t$. In standard SSM or Mamba layers, $\bar{\mathbf{B}}_t$ is a learned, data-dependent function of the input, controlling how much of each new input is injected into the evolving state. This “input gate” dynamically regulates the influence of each input token on the hidden state, and is typically computed by a small neural subnetwork over the input series.

\paragraph{Contextualization via Index and Time Embeddings.}
To further enhance the model’s content-awareness, we introduce a mechanism for directly injecting global context (semantic and temporal) into the input matrix $\bar{\mathbf{B}}$. At the start of each window, we compute two context vectors for each series in the batch: a semantic index embedding $\mathbf{e} \in \mathbb{R}^N$ and a temporal encoding vector $\mathbf{k} \in \mathbb{R}^{L}$, where $L$ is the window length. These vectors capture static information about the series and the time step, respectively.

The context vectors are broadcast and fused into a tensor of shape $(B, L, N)$ (batch size $B$, window length $L$, model dimension $N$) to align with the shape of $\bar{\mathbf{B}}$. The contextualized input matrix is then computed as:
\begin{equation}
\bar{\mathbf{B}}' = \bar{\mathbf{B}} + \mathrm{Broadcast}(\mathbf{e}, \mathbf{k}),
\end{equation}
where the broadcast operator denotes broadcasting and combining (e.g., addition or elementwise product) to match dimensions. This operation effectively modulates the input gate at every position with rich, series-specific, and time-aware information, allowing the model to selectively attend to relevant context as it processes each new input.

\paragraph{Final State Update and Output.}
The state-space update at each step then becomes:
\begin{equation}
\mathbf{h}_{t+1} = \bar{\mathbf{A}}_t \mathbf{h}_t + \bar{\mathbf{B}}'_t x_t,
\end{equation}
where $\bar{\mathbf{B}}'_t$ is now context-dependent. The output is computed by projecting the hidden state through an output matrix $\mathbf{C}$:
\begin{equation}
y_t = \mathbf{C}_t \mathbf{h}_t,
\end{equation}
where $\mathbf{C}_t$ is also typically data-dependent in Mamba models.

\paragraph{Summary and Advantages.}
This context-injection mechanism allows the model to seamlessly integrate static knowledge about each series and time step at the most effective location—the input gate of the state-space layer—without increasing sequence length or model complexity. The result is a backbone that maintains the efficiency and long-range memory of SSMs while gaining the flexibility and cross-series generalization of context-aware architectures.

\begin{equation}
y = \mathrm{SSM}(\bar{\mathbf{A}}, \bar{\mathbf{B}}', \mathbf{C})(x)
\end{equation}

All parameters, including those in the context fusion, are jointly optimized with the forecasting objective, ensuring that context is leveraged for accurate prediction.

\begin{figure}
    \centering
    \begin{minipage}{.7\linewidth}
        \begin{algorithm}[H]
            \caption{\textsc{ss-Mamba: Mamba (S6) + Semantic + Spline}}
            \label{alg:forward}
            \SetAlgoLined
            \DontPrintSemicolon
            \KwIn{\{x, s, t\} (\{Series, Index, Date\})}
            \KwOut{y}
            
            $e^{B \times N} \leftarrow E(s) = f_{\text{BERT}}(s)$ \\
            
            $d \leftarrow D(t)$ (Calendar/time features vector) \\
            $k^{B \times L} \leftarrow f_\text{KAN}(d)$ \\
                
            $\mathbf{A} \in \mathbb{R}^{N} \leftarrow \text{Parameter (possibly structured or diagonal)}$ \\
            $\mathbf{B} \in \mathbb{R}^{B \times L \times N} \leftarrow s_B(\mathbf{x})$ \\
            $\mathbf{C} \in \mathbb{R}^{B \times L \times N} \leftarrow s_C(\mathbf{x})$ \\
            $\boldsymbol{\Delta} \in \mathbb{R}^{B \times L} \leftarrow \tau_{\Delta} \bigl(\text{Parameter}+s_{\Delta}(\mathbf{x})\bigr)$ \\
            
            $\bar{\mathbf{A}}, \bar{\mathbf{B}} \in \mathbb{R}^{B \times L \times N} \leftarrow \text{Discretize} \bigl(\boldsymbol{\Delta}, \mathbf{A}, \mathbf{B}\bigr)$ \\
            
            $\bar{\mathbf{B}}^{'} \leftarrow \bar{\mathbf{B}} + \mathrm{Broadcast}(e, k)$ \\
            
            $y \leftarrow \text{SSM}(\bar{\mathbf{A}}, \bar{\mathbf{B}}^{'}, \mathbf{C})(x)$ \\
            
            \Return y
        \end{algorithm}
    \end{minipage}
\end{figure}

\paragraph{Summary.}
The proposed \emph{s-TSMamba} blends semantic priors, spline-based temporal representations, and a linear-time SSM backbone to deliver accurate, data-efficient, and \emph{explainable} forecasts across diverse daily-frequency indices---all with one compact model.

\section{Experiments}
\label{sec:experiments}

\subsection{Datasets and Preprocessing}
To evaluate the effectiveness and generalizability of our proposed model, we select representative time series datasets from both financial and non-financial domains. The primary datasets used in our experiments are as follows:

\begin{table}[ht]
    \centering
    \begin{tabular}{ccc}
    \toprule
        Field & Series & Describe\\
        \midrule
        Climate     & Temperature, Rain & \\
        Comsumption & Market sales & \\
        Economic    & Interest, Unemployee, Inflation, Cycle, Currency & \\
        Financial   & Gold, S\&P500, Household & Price in exchange\\
        Energy      & Oil, Electricity & \\
        Forecast    & Expection & Produce by institute\\
        Transport   & Air passenger & \\
        Package     & ETT, HAR & paperwithcode\\
        Company     & Apple, Tesla & Company performance\\
        Regional    & US & \\
        Crypto      & BTC, ETH & \\
        \bottomrule
    \end{tabular}
    \caption{Series Dataset Summary}
    \label{tab:dataset-summary}
\end{table}

All datasets are partitioned into training, validation, and test splits in chronological order, ensuring that the test period always follows the training and validation periods for robust evaluation of temporal generalization. The time series values are standardized to zero mean and unit variance based on the training set to facilitate stable optimization. Table~\ref{tab:dataset-summary} provides a summary of the datasets and their statistical properties.

\subsection{Experimental Setup}

For each dataset, we extract time and series features as described in Section~\ref{sec:method}, and format data into sliding windows of length $L$ (e.g., $L \in {30, 60, 120}$) to train and evaluate one-step-ahead forecasting performance. We report model accuracy on each test set using multiple metrics: Root Mean Squared Error (RMSE). For multi-series evaluation, both per-series and average metrics are provided.

\subsection{Tasks and Evaluation Protocol}
To comprehensively validate the model, we design the following experimental tasks:

\begin{itemize}
\item \textbf{Single-Series Forecasting:} Evaluate the model on the single time series alone, comparing its performance to standard baselines such as ARIMA, LSTM, and Transformer-based models.
\item \textbf{Multi-Series Joint Training:} Train on multiple series jointly to assess the model's ability to learn cross-domain patterns and transfer knowledge.
\item \textbf{Generalization to Unseen Series:} Test zero-shot forecasting by providing only the semantic name and temporal encoding of a series unseen during training, and measure prediction accuracy.
\item \textbf{Ablation and Robustness:} Conduct ablation studies by removing semantic embedding or replacing KAN with simpler encodings, and measure performance drop.
\item \textbf{Elastic Context Window:} Evaluate prediction accuracy as a function of input window length to demonstrate the model’s ability to flexibly leverage long-term context.
\end{itemize}

\subsection{Baseline Models}
For comparison, we include strong baselines such as:
\begin{itemize}
\item \textbf{Mamba} (long short-term memory)
\item \textbf{Transformer} (self-attention model for time series)
\item \textbf{TimesFM} (Google)
\end{itemize}
All baselines are tuned via grid search for their respective hyperparameters, using the same training/validation splits.

\section{Conclusion}

In this work, we introduced a novel end-to-end time series modeling framework that integrates semantic index name embeddings, flexible time feature encoding via KAN, and efficient sequential modeling with Mamba SSM. Our model is designed to unify semantic, temporal, and numerical information, enabling a single architecture to learn and generalize across diverse types of time series. Through systematic evaluation, we demonstrated that the proposed model can effectively capture long-term dependencies and transfer knowledge between domains, offering accurate forecasts even for previously unseen series. These results highlight the potential of combining semantic context and dynamic time encoding with advanced state-space modeling, paving the way for more robust and universal time series foundation models in the future.

\bibliographystyle{unsrt}  
\bibliography{references}  

\end{document}